\documentclass[final,custom]{anthology-ch-arxiv}
\usepackage{booktabs}
\usepackage{graphicx}
\usepackage{subcaption}
\usepackage{wrapfig}

\usepackage{longtable}
\title{Measuring Gender Representation in Animated Films}
\author[1]{David Bamman}[
  orcid=
]
\author[2]{Allison Cooper}[
  orcid=
]
\author[1]{Ruby Alvarez Rubio}[
  orcid=
]
\authorlinebreak
\author[1]{Reina Kushihashi}[
  orcid=
]
\author[1]{Madison Mar}[
  orcid=
]
\affiliation{1}{UC Berkeley}
\affiliation{2}{Bowdoin College}
\keywords[eng]{animated films, gender representation, cultural analytics}
\customhead{}
\makeatletter
\renewcommand{\@firstpagefootnote}{\footnotesize \textcolor{newgray}{© 2026 by the authors. Licensed under \href{https://creativecommons.org/licenses/by-sa/4.0}{Creative Commons Attribution-ShareAlike 4.0 International (CC BY-SA 4.0).} Frame captures from the films discussed are included under fair use for the purposes of criticism and scholarship. Copyright in these images belongs to their respective owners.}}
\makeatother
\begin{document}

\maketitle

\begin{abstract}
Animated films---often developed with an audience of children in mind---are an important vector for enculturation, and empirical work that has examined the representation of gender at scale in these films has largely focused on counting the gender composition of the cast rather than deploying a more fine-grained instrument (such as assessing the visibility of those characters in overall screentime).  In this work, we develop a computational pipeline for recognizing animated characters in these films, and use it to test several hypotheses about gender representation in a corpus of 224 popular animated movies.  We find that while the overall representation of female characters in animated films largely tracks with those of live-action films (over the period 1980--2025), we see stark differences between the representation of human characters (much greater representation among women and girls) and non-humans (largely male).  Contrary to past work on Disney, we do not see female characters declining in antagonist roles in animated films, and characters who are women and girls are much more likely to share scenes together than their live action contemporaneous counterparts.
\end{abstract}

\section{Introduction}

\begin{figure}[t]
\centering

\begin{subfigure}{0.325\textwidth}
  \includegraphics[width=\linewidth]{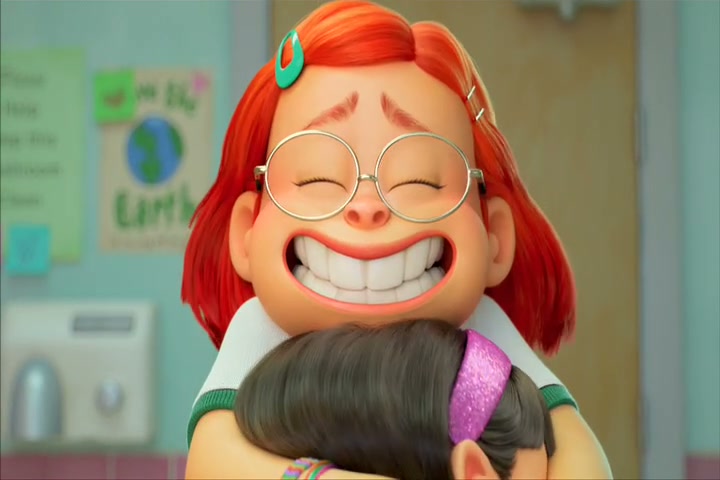}
\end{subfigure}
\begin{subfigure}{0.325\textwidth}
  \includegraphics[width=\linewidth]{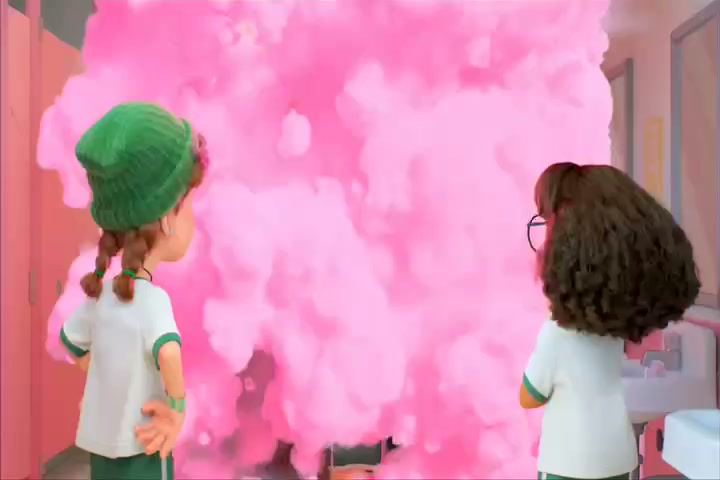}
\end{subfigure}
\begin{subfigure}{0.325\textwidth}
  \includegraphics[width=\linewidth]{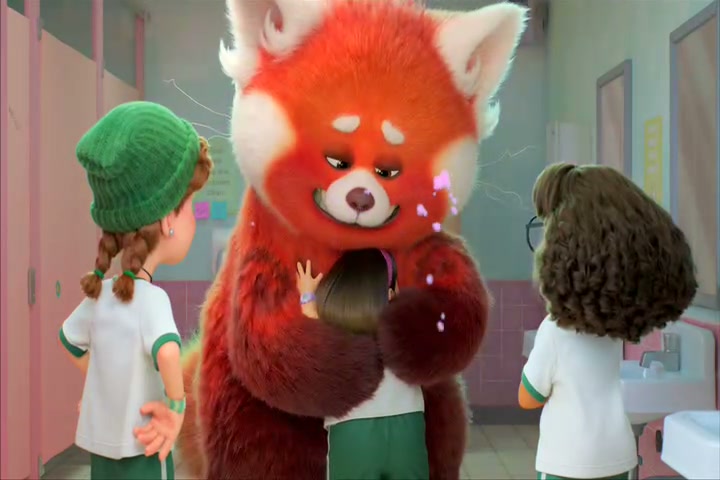}
\end{subfigure}

\caption{\label{turningred}\emph{Turning Red}.}
\end{figure}

In the 1940s, filmmaker and theorist Sergei Eisenstein observed that the appeal of Walt Disney’s animation lay in its ability to assume any form whatsoever\ \cite{EisensteinSergei1988EoD}. Disney’s work was irresistibly attractive, Eisenstein mused, “in a country and social order with such a mercilessly standardized and mechanically measured existence” (5). Nearly a century later, a film like \textit{Turning Red }(Domee Shi, 2022) confirms animation's continuing capacity for fluidity and metamorphosis, transforming its 13-year-old protagonist Mei into a red panda whenever she experiences the turbulent emotions that accompany adolescence (fig. \ref{turningred}). \textit{Turning Red} exemplifies Eisenstein’s ideas about animation and shows what makes it distinct from live-action: its frames are created rather than captured and, free from the constraints of the pro-filmic world required by live-action, director Shi can imagine a world where traditional binaries like male-female and animal-human are easily overturned, providing in their place a complex and sensitive exploration of individual identity. 

As \textcite{dobson2018animation} argue, this kind of imaginative visualization sets animation apart from live action, offering unique opportunities for formal experimentation and expressive freedom. A central objective of our work is to understand how animated films have exercised that expressive freedom in relation to gender representation and how they compare in this regard to their live action counterparts. As a wholly created world, animation need not inherit the same biases as live action filmmaking.  But does it still?

One of the most persistent findings in studies of gender in popular culture is the continued overrepresentation of men (and male non-human characters) across just about every medium. We see this pattern not only across 43 years of live-action Hollywood films\ \cite{bammanpnas}, but also in other equally influential forms of storytelling, including English-language fiction\ \cite{underwood2018transformation} and children's books\ \cite{casey2021sixty,walsh2025bears}. Across all of these mediums, male characters appear roughly three times more often than female characters. These representational inequalities are important in assessing how children learn about gender, and this has motivated much work on children's literature in the past\ \cite{horst2015open,abad2013storybooks}.  Indeed, structuralist theory (\cite{althusser2006lenin}) and scholarship from a range of areas like critical pedagogy (\cite{giroux2001mouse}), communication studies (\cite{gerbner2002growing}), sociology (\cite{tuchman1978hearth}), and developmental psychology (\cite{bandura1986social}) has converged around the idea that patterns of representation in popular media are far from incidental: they carry ideological weight and developmental consequences. Advancing quantitative methods for the analysis of these kinds of patterns in moving image media is increasingly urgent, we believe, given the steady shift away from page to screen in American society\ \cite{bls_atus_2024}.  

Past computational work has developed  pipelines for measuring such representation in live-action films\ \cite{bammanpnas}, relying on the maturity of computer vision models for reasoning about the human form. In this work, we build on a related line of research recognizing characters in still-image cartoons\ \cite{zheng2020cartoon} and animation\ \cite{gui2025character}, training and evaluating models for character detection and recognition, manually building cast lists for movies and linking gender information (from voice actors and fan communities) to each character in order to support this analysis. Our development of these computational methods allows us to meaningfully expand upon existing empirical approaches that have examined gender representation in animated films primarily by counting the gender makeup of characters in the cast\ \cite{raju2023cohort,geena2019benchmark,geena2019seejane,teran2024seejane}, extending that work to incorporate more fine-grained measures. How many minutes is each of those cast members visible on screen?  How do they interact with each other? What role do they play?
\\[10pt]
\noindent
In applying this computational lens to animated films, our work makes three contributions:
\begin{itemize}
    \item We train and evaluate a pipeline for recognizing characters in animated films, and release models for others to use on their own data.
    \item We apply this pipeline to a collection of 224 popular films to test several hypotheses about the representation of gender in animated films, placing those findings in conversation with representation in contemporaneous live-action films.
    \item We release data produced by this pipeline for others to explore, including the location of all detected characters in all films, information about character gender, and role as protagonists/antagonists.
\end{itemize}

This work reveals several important findings. Despite representing entirely created worlds, animation looks to recreate the same patterns of gendered screentime as live-action ones, but this first glance is complicated by the greater space of characterization that animation opens up. Women and girls (i.e., human characters), for example, have much greater screentime than their live-action counterparts, but non-human characters (e.g., animals) are even more strongly skewed masculine---revealing  androcentrism's enduring nature \cite{bem1993lenses} notwithstanding the apparent lack of real-world pressures to encode gender this way.  At the same time, female characters in animation are increasingly sharing the screen together, suggesting that, in contrast to the comparative isolation of early protagonists in \emph{Cinderella} and \textit{Sleeping Beauty}, they are gaining a narrative social existence independent of men. Finally, while live-action movies have seen a trend distancing women as strong antagonists, this association has continued in animation, granting them access to some of the medium's most narratively rich characters (e.g. Te Kā in \textit{Moana}). Overall, this work sheds light on the ways in which animation both exploits the fluidity of form that Eisenstein alludes to, while also retaining the same biases of the world in which it is made.

\section{Data}\label{sec:data}

We focus in this work on popular films, drawing data on historical box office numbers from two sources: for the period 1980 to the present, we use calendar grosses from Box Office Mojo.\footnote{\url{https://www.boxofficemojo.com/}} For the period 1922-1979, we draw on a dataset of weekly historical box office numbers from \emph{Variety} magazine released by \textcite{bamman2026evaluating}.  Both sources measure the amount of money spent on a film in a given calendar year.  We select animated films in the top 50 movies by box office each year, purchase all that we are able to acquire, and digitize them under the affordances of 37 CFR 201.40(b)(4), an exemption to the US Digital Millennium Copyright Act for text and data mining research.  While box office grosses capture of course only one measure of significance, it provides a direct assessment of the overall reach of a film (providing a proxy to the number of people who have viewed it, and thereby the potential influence of its representation of gender).

\begin{figure}[htbp]
    \centering
    \includegraphics[width=.8\linewidth]{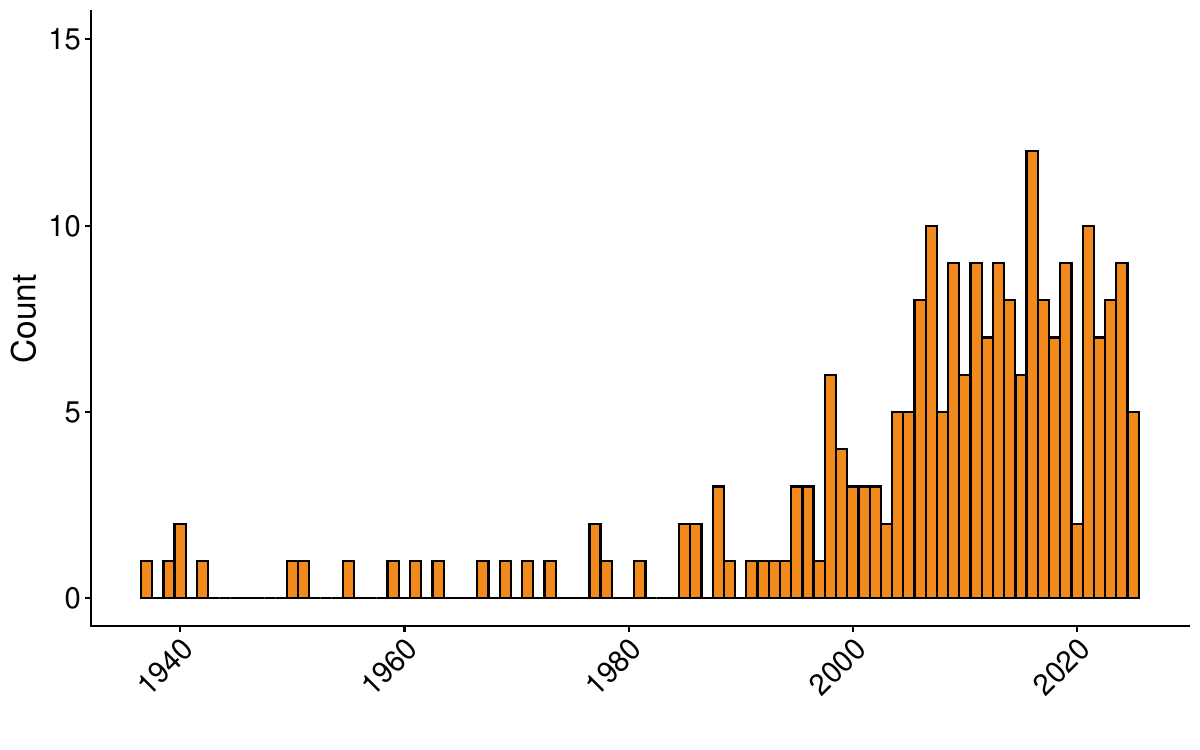}
    \caption{Distribution of animated movies in our collection over time.}
    \label{fig:movietime}
\end{figure}

In total, our collection includes 224 movies, spanning 1937 (\emph{Snow White and the Seven Dwarfs}) to 2025, with 78\% of it released after the year 2000 (as illustrated in fig. \ref{fig:movietime}); 2016 alone saw 12 of the 50 highest-grossing movies being animated.  While there is a long tail of theatrical releases each year,\footnote{Wikipedia's \emph{List of animated feature films released theatrically in the United States} enumerates approximately 1,000 from 1937 to today.} the movies in our collection comprise 66.7\% of the total box office grosses for all animated films documented in Box Office Mojo or \emph{Variety}.

The collection is dominated by major commercial animation studios, including 59 movies from Disney (\emph{Pinocchio, The Lion King, Zootopia}), 46 movies from DreamWorks (\emph{Shrek, Kung Fu Panda, The Bad Guys}), 26 movies from Pixar (\emph{Toy Story, The Incredibles, Inside Out}), and 15 movies from Illumination (\emph{Despicable Me, Sing}), among others. An industrial shift toward franchising over the last 50 years has affected animated films as much as live action. This is reflected in our corpus, with nearly 47\% of the films it contains belonging to franchises that include at least two movies in the top 50 by box office (e.g. \textit{Toy Story 1-4}).  \textcite{loock2024hollywood} argues that Hollywood remakes narratives for a variety of reasons, with financial risk-management perhaps the most obvious economic one. But she also details the important cultural reasons for remakes, which update narratives to reflect social changes, meeting each new generation in its own era. This latter point is especially relevant for our study, which generally documents increasing screentime in the \#MeToo era for female characters across franchises.

\section{Methods}

In order to measure screentime for animated characters, we need a measuring instrument tailored to that domain.  In order to enable direct comparison with live-action films, we build primarily on the character recognition pipeline of \textcite{bammanpnas}: we segment a movie into shots using TransNetV2\ \cite{soucek2020transnetv2}, detect the character faces that are present in each frame (\S\ref{sec:detection}), assemble overlapping faces in sequential frames into \emph{face tracks}\ \cite{bochinski2017high}, generate a representation for the single highest-confidence face in each track (\S\ref{sec:recognition}) and find its closest match among the cast list for the film (\S\ref{castlists}), from which we derive the character's gender (\S\ref{sec:chargender}).  Each of these steps presents challenges for animated film, which we detail below.

\subsection{Detection}\label{sec:detection}

Face detection for human faces has made significant progress over the past decade, with state-of-the-art systems achieving accuracies (in average precision) for difficult benchmarks like WIDER Face-Hard of 92\%\ \cite{deng2020retinaface,zhu2020tinaface}; within the same domain of movies, \textcite{bammanpnas} report an AP@50 of 86.9. This work benefits from the existence of large training datasets for human faces, and the constraints imposed by the human form; characters in animated movies exhibit a much wider range of appearance, however, including human-like characters (\emph{Frozen, The Simpsons}), animals (\emph{Zootopia}, \emph{Madagascar}), machines (\emph{Cars, Planes}), and fantastic creatures (e.g., \emph{Monsters, Inc.}).

To measure the degree to which we can detect these difficult characters, we draw on two datasets: first, iCartoonFace\ \cite{zheng2020cartoon}, which annotates 60,000 images (50K training, 10K test) of cartoon characters drawn from lists originating in Wikipedia, with images sourced from the public web.  We use the given splits for training and evaluation.  To assess the degree to which we can recognize characters in the wild within the domain of film itself, we also create an evaluation set of 2500 frames (25 images drawn from 100 animated films).  For each frame, we manually label the bounding box around each character's observed face, identifying a total of 3700 faces.

We use this data to train and evaluate several classes of detectors: RetinaFace\ \cite{deng2020retinaface}, a ResNet-based face detector pretrained on WIDER Face; YOLO26-L\ \cite{jocher2026ultralytics}, a CNN-based one-stage object detector pretrained on MS-COCO\ \cite{lin2014microsoft}; RT-DETR\ \cite{lv2024rtdetrv2improvedbaselinebagoffreebies}, a transformer-based object detector also pretrained on MS-COCO; YuNet-N\ \cite{wu2023yunet}, a small CNN-based face detector pretrained on WIDER Face; and YOLOE-26L\ \cite{sapkota2026yoloe}, an open-vocabulary YOLO detector.  For all trainable models, we apply a common set of image augmentations during training (random square crop, horizontal flip, color jitter, mosaic augmentation\ \cite{bochkovskiy2020yolov4}), and for all models, we separately evaluate a test-time augmentation (\texttt{+TTA}) strategy, which resizes the longest image side to \{640, 1100, 1600\} pixels and applies a horizontal flip, yielding 6 variants for every image; the same detector runs on all 6 variants and detected faces are merged back into the original image dimensions, with non-maximum suppression (NMS) to remove overlapping boxes.

All trainable models are trained on three data sources: training exclusively on WIDER Face\ \cite{yang2016wider}, which captures only real-life, human faces (to get a sense of the performance of out-of-the-box human face-detection models); training on iCartoonFace alone; and training on a mixture of WIDER Face and iCartoonFace (since prior work has found that mixture helpful\ \cite{zheng2020cartoon}).  We use the iCartoonFace training data to select the best model, splitting the 50K images in the original split into 45k training and 5k development; we train all models for 40 epochs, and the epoch with the best AP@50 score on that development data is selected as the best model to evaluate on the test data.

Table \ref{tab:filmdetect} reports the results on the 100 annotated films, along with 95\% confidence intervals (block resampling at level of films to provide the most conservative estimate); table \ref{tab:icartoon_test} (appendix \ref{appdx:detect}) illustrates results on the held-out iCartoonFace test data. Several findings emerge: first, models trained only on human faces struggle to detect animated ones (necessitating the current study); the transformer-based RT-DETR model is the best performing, eclipsing the closest CNN-based model (YOLO26-L) by several points; test-time augmentation generally improves the lower end of performance, but has diminishing returns as models improve and comes at a significant computational cost during prediction.  To balance the overall accuracy of the model with its computational speed, we select the RT-DETR-L model with no TTA for the remaining experiments.

\begin{table}
\centering
\begin{tabular}{lrrrr}
\toprule
Model & Train WF & Train ICF & Train WF+ICF & im/s \\
\midrule
RetinaFace-R50 & 32.11 {\scriptsize [27.08, 37.11]} & 78.73 {\scriptsize [75.62, 81.75]} & 78.49 {\scriptsize [75.50, 81.43]} & 57.5 \\
\quad + TTA & 38.77 {\scriptsize [33.88, 44.54]} & 80.68 {\scriptsize [78.18, 83.42]} & 81.69 {\scriptsize [79.07, 84.31]} & 4.0 \\
YOLO26-L & 39.72 {\scriptsize [34.71, 45.07]} & 80.81 {\scriptsize [77.98, 83.56]} & 80.85 {\scriptsize [78.42, 83.64]} & 91.4 \\
\quad + TTA & 45.40 {\scriptsize [40.69, 50.51]} & 83.10 {\scriptsize [80.33, 85.48]} & 83.43 {\scriptsize [81.13, 85.87]} & 13.8 \\
RT-DETR-L & 40.76 {\scriptsize [36.11, 45.94]} & {\color{magenta}83.49} {\scriptsize [80.99, 85.91]} & 82.83 {\scriptsize [80.10, 85.25]} & 71.6 \\
\quad + TTA & 43.14 {\scriptsize [38.41, 47.85]} & 84.21 {\scriptsize [81.52, 86.49]} & 83.50 {\scriptsize [81.14, 85.87]} & 8.2 \\
YuNet-N & 37.42 {\scriptsize [32.37, 42.24]} & 66.12 {\scriptsize [62.56, 69.71]} & 66.51 {\scriptsize [62.58, 70.17]} & 140.4 \\
\quad + TTA & 42.16 {\scriptsize [37.53, 47.04]} & 65.09 {\scriptsize [61.33, 68.64]} & 66.58 {\scriptsize [63.32, 70.11]} & 23.5 \\
\midrule
YOLOE-26L (text-prompted) & \multicolumn{3}{c}{16.99 {\scriptsize [15.09, 19.24]}} & 49.2 \\
\quad + TTA & \multicolumn{3}{c}{24.72 {\scriptsize [22.59, 27.37]}} & 7.4 \\
\bottomrule
\end{tabular}
\caption{
AP@0.5 (95\% bootstrap CI) on film data (per-movie), plus inference throughput (images/second). We select the highlighted model for subsequent experiments.}
\label{tab:filmdetect}
\end{table}

\subsection{Recognition}\label{sec:recognition}

Likewise, the task of face recognition---identifying \emph{whose} face is shown in an image---has benefited from large datasets for training and a variety of computational models and loss functions optimized for the human form.  iCartoonFace again provides a source for training and evaluation on this task, containing nearly 400k images of 5,013 cartoon identities. As with detection, we  create an evaluation dataset for film as well, manually grouping character faces belonging to the same identity in our collection of 100 films.

We train and evaluate two classes of models on this data: the \texttt{buffalo\_l} model from Insightface\ \cite{guo2019insightface}, a ResNet-50 model trained with an ArcFace loss function on the WebFace600K dataset; and DINOv2\ \cite{oquab2023dinov2}, a vision transformer trained for general visual representation in a self-supervised manner (not optimized specifically for face recognition); we evaluate both the base model (\texttt{vitb14}) and large (\texttt{vitl14}).  We use Rank@1 as an evaluation metric: for iCartoonFace, we follow their implementation and calculate the cosine similarity between a source image and a single reference image of the same identity + 2500 distractor identities; Rank@1 measures the degree to which the most-similar image by cosine similarity among these 2501 images is the single image with the same identity as the source.  For our film dataset, we draw distractors only from the same movie, which mirrors the way in which this method will be used in analysis (identifying the closest reference image among the cast list for a movie); source images come from any identity with at least two faces in the evaluation data; all other faces of alternative identities (including those only observed once) are used as distractors. 

We assess four different training scenarios: the \emph{base} scenario simply runs each model out-of-the-box to generate a representation for each image (i.e., without training on iCartoonFace), while the \emph{FT} scenario fine-tunes on iCartoonFace using an ArcFace loss\ \cite{deng2019arcface}.  We again use the iCartoonFace training data to find the best model, using 10\% of identities in the training portion as development data and selecting the checkpoint that maximizes the mean reciprocal rank (for every probe image, the mean reciprocal of the rank assigned to the true match among the distractors).

But film data presents another natural form of self-supervision for character recognition: the face tracks themselves.  Face tracks hold visual continuity in successive frames; as \textcite{bochinski2017high} note, at high frame rates (e.g., 24 fps), faces move little from one frame to the next, and so a simple intersection-over-union measure in pixel overlap between successive frames is sufficient to accurately track a face moving in time. For long tracks, the starting position of a face and its ending position within the track may have very different appearances and orientations; while not as strong of a signal as identities that are observed in completely different contexts, it does provide a natural form of self-supervision that can be useful for the problem of domain adaptation (moving from cartoon images to animated films).  In this form of training, we train with a triplet loss, selecting a positive pair as two faces from the same face track separated by at least one second; a negative example is drawn from the same \emph{frame} as a positive example (so we can be relatively sure that they belong to different identities).  

We explore this self-supervision in two scenarios.  First, by training only on movies that do not occur in the test data (\texttt{I+T/Train}); this provides a measure of adaptation to the form of Hollywood cinema, and provides the closest measure of generalization performance.  Second, by training on all movies (\texttt{I+T}), which provides a measure of adaptation to the specific characters in our object of study. In neither case does this training require labeled data of any kind. \texttt{I+T} can be read as expected future recognition performance if allowed to adapt to the films themselves; \texttt{I+T/Train} is the future recognition performance if this domain adaptation is not possible (i.e., a scenario where a researcher runs this trained model on a new movie without performing further adaptation to it).

For all models, we explore two options for crop size: a close crop, which generates a representation for a face based on its exact bounding box; and a slightly wider crop that adds 25\% to each dimension, capturing some information (such as hair and some clothes) that may be useful for representation. Models in this training regime are initialized with their corresponding FT-trained model (to capture the performance learned from iCartoonFace).

The results of all models can be found in table \ref{tab:recognition_variants_film}, along with 95\% confidence intervals (block resampling again at level of films); results for iCartoonFace test data can be found in table \ref{tab:recognition_variants_icartoonface} (appendix \ref{appdx:recog}), and mirrors the reported performance of that paper.  Several observations again arise: 25\% padding on the crop size generally leads to major improvements across models, echoing previous findings\ \cite{gui2025character}; the DINOv2 vision transformer far exceeds the ResNet-based \texttt{buffalo\_l} model on this task; increasing model size within the DINO family also leads to general improvements, but comes at a significant computational cost.  Fine-tuning on the tracks to provide a form of domain adaptation (both to cinema and to the specific movies/characters) also yields improvements, and especially so for the larger \texttt{vitl14} model.  Given this high accuracy, we select the DINO \texttt{vitl14} model for experiments for the rest of this work, despite its computational cost (though note that we run this model only on one face per face track, not once per frame, so its cost is more constrained compared to a face detection model).

\begin{table}
\centering
\resizebox{\linewidth}{!}{%
\begin{tabular}{lccccr}
\toprule
Model & Base & FT & I+T/Train & I+T & im/s \\
\midrule
buffalo\_l & 0.2246 {\tiny [0.1986, 0.2523]} & 0.4485 {\tiny [0.4060, 0.4915]} & 0.4653 {\tiny [0.4204, 0.5111]} & 0.4819 {\tiny [0.4354, 0.5285]} & 312.9 \\
\quad + 25\% padding & 0.2684 {\tiny [0.2366, 0.3003]} & 0.5138 {\tiny [0.4720, 0.5553]} & 0.5434 {\tiny [0.5009, 0.5893]} & 0.5617 {\tiny [0.5188, 0.6043]} & 312.9 \\
DINOv2 vitb14 & 0.3085 {\tiny [0.2717, 0.3462]} & 0.5904 {\tiny [0.5418, 0.6378]} & 0.5956 {\tiny [0.5452, 0.6432]} & 0.6160 {\tiny [0.5681, 0.6599]} & 463.2 \\
\quad + 25\% padding & 0.2718 {\tiny [0.2344, 0.3099]} & 0.6785 {\tiny [0.6355, 0.7209]} & 0.7130 {\tiny [0.6741, 0.7530]} & 0.7160 {\tiny [0.6825, 0.7505]} & 463.2 \\
DINOv2 vitl14 & 0.3443 {\tiny [0.3048, 0.3853]} & 0.6329 {\tiny [0.5883, 0.6770]} & 0.6426 {\tiny [0.5943, 0.6902]} & 0.6986 {\tiny [0.6533, 0.7410]} & 155.1 \\
\quad + 25\% padding & 0.2873 {\tiny [0.2497, 0.3302]} & 0.7079 {\tiny [0.6690, 0.7484]} & 0.7176 {\tiny [0.6815, 0.7579]} & {\color{magenta}0.7711} {\tiny [0.7385, 0.8036]} & 155.1 \\
\bottomrule
\end{tabular}
}
\caption{Rank@1 identification accuracy (95\% bootstrap CI) on film data (per-movie), crop 0\% and crop 25\% (indented), plus inference throughput (images/second). We select the highlighted model for subsequent experiments. }
\label{tab:recognition_variants_film}
\end{table}

\subsection{Building a cast}\label{castlists}

Prior work measuring actor screentime in film was able to do so by relying on the existence of cast lists, matching visual representations of characters on screen with equivalent representations of actors from cast photos on IMDb.  As \textcite{gui2025character} note, such an exhaustive resource does not exist for animated films and presents one of the core challenges to this work; cast lists on IMDb for animated films cover the voice actors---leaving absent any non-speaking characters---and largely provide visual representations of the human actors, and not the animated characters they voice.  

We create casts for all movies in our collection through a human-in-the-loop clustering process, drawing on partial cast information from two sources: \texttt{www.behindthevoiceactors.com} (which notes the voice actors behind speaking roles, along with images of the character); and \texttt{www.fandom.com}, which presents character information for a movie universe (e.g., all \emph{Toy Story} movies).  We run our choice of face-detection model above (RT-DETR-L) on all frames of all movies, assemble them into face tracks, and generate a representation for the single highest-confidence face in each track using our choice of face-recognition model (DINOv2 \texttt{vitl14} with 25\% padding).  We run the same process on any character images from those sources, treating each source image as a seed for an entity cluster and finding the twelve nearest neighbor faces in the movie to each seed image.  In order to ensure that we capture non-speaking characters as well, we also run any remaining images (i.e., that are not among the nearest neighbors to any seed) through k-means clustering (setting $k=50$), and display the twelve images closest to each cluster centroid.  Annotators (all co-authors) then use this starting point to manually create a cast list, where characters are paired with images from within the movie that represent them.  Annotators are free to merge multiple existing clusters and remove individual images that do not reflect the character.  Appendix \ref{appdx:clustering} displays the clustering interface with the 1939 movie \emph{Gulliver's Travels}. We carry out this process for all 224 movies in our collection, and release the cast lists (along with DINOv2 vector representations for each of them) as part of this work.

\begin{figure}[htbp]
    \centering
    \includegraphics[width=.8\linewidth]{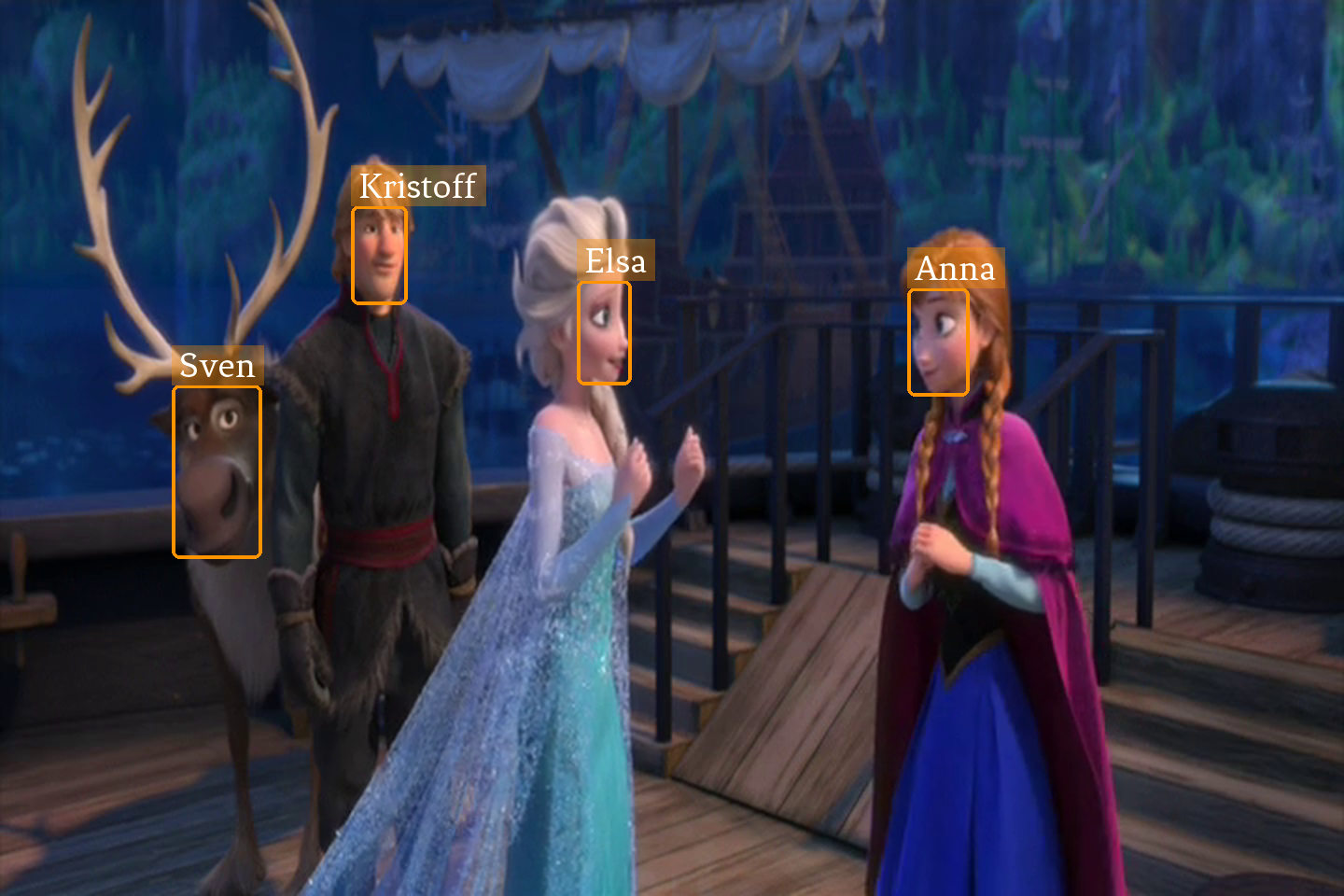}
    \caption{Output of computational process.}
    \label{fig:output}
\end{figure}

With these visual representations of the cast, we can then complete the process of identifying the characters who are present on screen.  We create an average vector representation of each character in the cast from all images in the movie that an annotator has assigned to them; we then match every face detected in the film to this candidate character bank, treating as a match any face with a cosine similarity higher than 0.42 to its closest candidate (optimized on annotated data).  Figure \ref{fig:output} illustrates this output for one frame from the 2013 movie \emph{Frozen}; this process allows us to match speaking characters such as Elsa, Anna and Kristoff, along with non-speaking characters (such as the reindeer Sven).

\subsection{Character gender}\label{sec:chargender}

For all analyses that follow, we focus on characters that have speaking roles in the film (i.e., that are voiced by human actors), matching character names to  actors using cast lists provided by \url{www.behindthevoiceactors.com} and IMDb.

We rely on two sources of information to capture character gender. First, we look to the gender of the voice actor (\textsc{Voice actor gender}), as reflected in Wikidata; this captures the gender perception of the Wikimedia community (with a range of gender values beyond a simple binary), often with sourcing to attestations online.  This gender information is partial: not all actors have gender information recorded on Wikidata, and several characters are voiced by a human actor whose gender does not align with that coded for the character---e.g., Bart Simpson in \emph{The Simpsons Movie} is voiced by a woman (Nancy Cartwright), while Bart is referred to as a boy in the film. To fill in gaps and identify such cases of misalignment, we draw on the community wikis of \url{www.fandom.com}, which contain elaborate descriptions of nearly all films (96.9\%) in our collection, including detailed descriptions of characters.  We match characters from the film to characters in the fandom domain for each film, and measure the community perception of each character's gender (\textsc{Community gender}) through their relative use of \emph{he/she} pronouns on the page\ \cite{reagle2011gender} (which past work has shown to be extremely accurate for measuring gender on Wikipedia\ \cite{bamman2014unsupervised}) and by explicit mentions of the character being non-binary. Across all characters in our collection, only one is described as non-binary by the fandom community: the character Lake Ripple from the 2023 movie \emph{Elemental}. Strategies like queer-coding date back to the origins of film itself and can offer at least an oblique measure of the representation of non normative sexuality \cite{piluso2023above}, but overt self-identification as non-binary is a recent enough phenomenon that we are not necessarily surprised by this finding in relation to gender. 

For all other characters, we calibrate \textsc{Community gender} on the strength of evidence (the number of observed pronouns), by adopting a Bayesian perspective. For each character with $c_{she}$ feminine pronoun mentions and $c_{he}$ masculine pronoun mentions, we model the probability that a character is female as $\theta$. Using a Beta-Binomial model with prior $\theta \sim \textrm{Beta}(\alpha,\beta)$, the posterior distribution is  $\theta \mid c_{she}, c_{he} \sim \textrm{Beta}(\alpha+c_{she}, \beta+c_{he})$, where we set $\alpha=\beta=5$ to encode a weakly informative prior that does not favor either gender while requiring sufficient pronoun evidence to push our belief beyond ``unknown.'' We count a character as female if $P(\theta > 0.50 \mid c_{she},c_{he}) > 0.95$, and as male if $P(\theta < 0.50 \mid c_{she},c_{he}) > 0.95$. Characters that do not meet either criterion are left unclassified.  We use this information directly for any character without voice actor gender information, and manually review any cases where \textsc{Voice actor gender} and \textsc{Community gender} disagree.

\section{Analysis}

The pipeline described above allows us to measure the screentime for characters in animated films and attribute their gender in order to measure the changing dynamics of gender representation on screen in our collection.  This directly allows us to answer our primary research questions.

\subsection{Gender representation}

\begin{figure}[t]
  \centering
  \begin{subfigure}[b]{0.48\textwidth}
    \includegraphics[width=\textwidth]{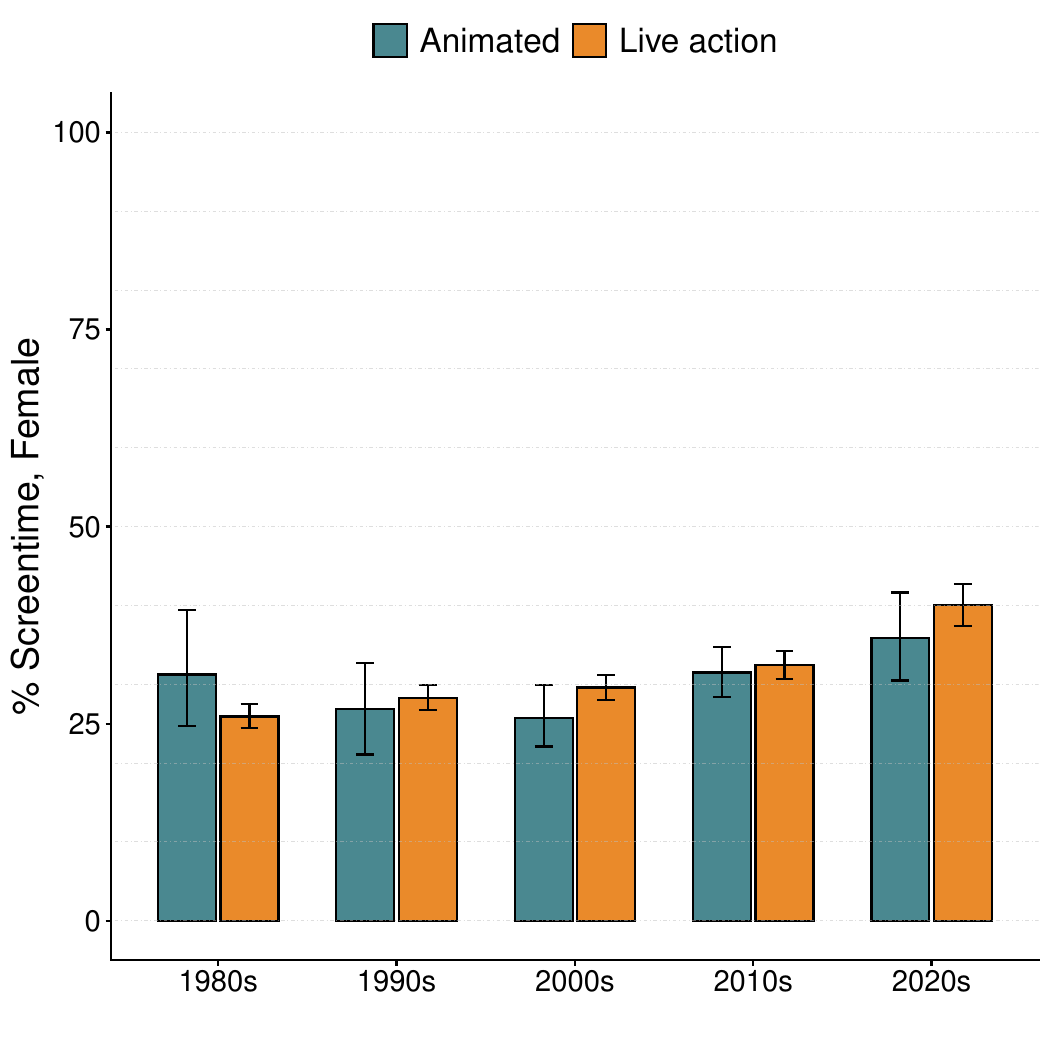}
    \caption{Total animated vs.\ live action}
    \label{fig:total-vs-live-action}
  \end{subfigure}
  \hfill
  \begin{subfigure}[b]{0.48\textwidth}
    \includegraphics[width=\textwidth]{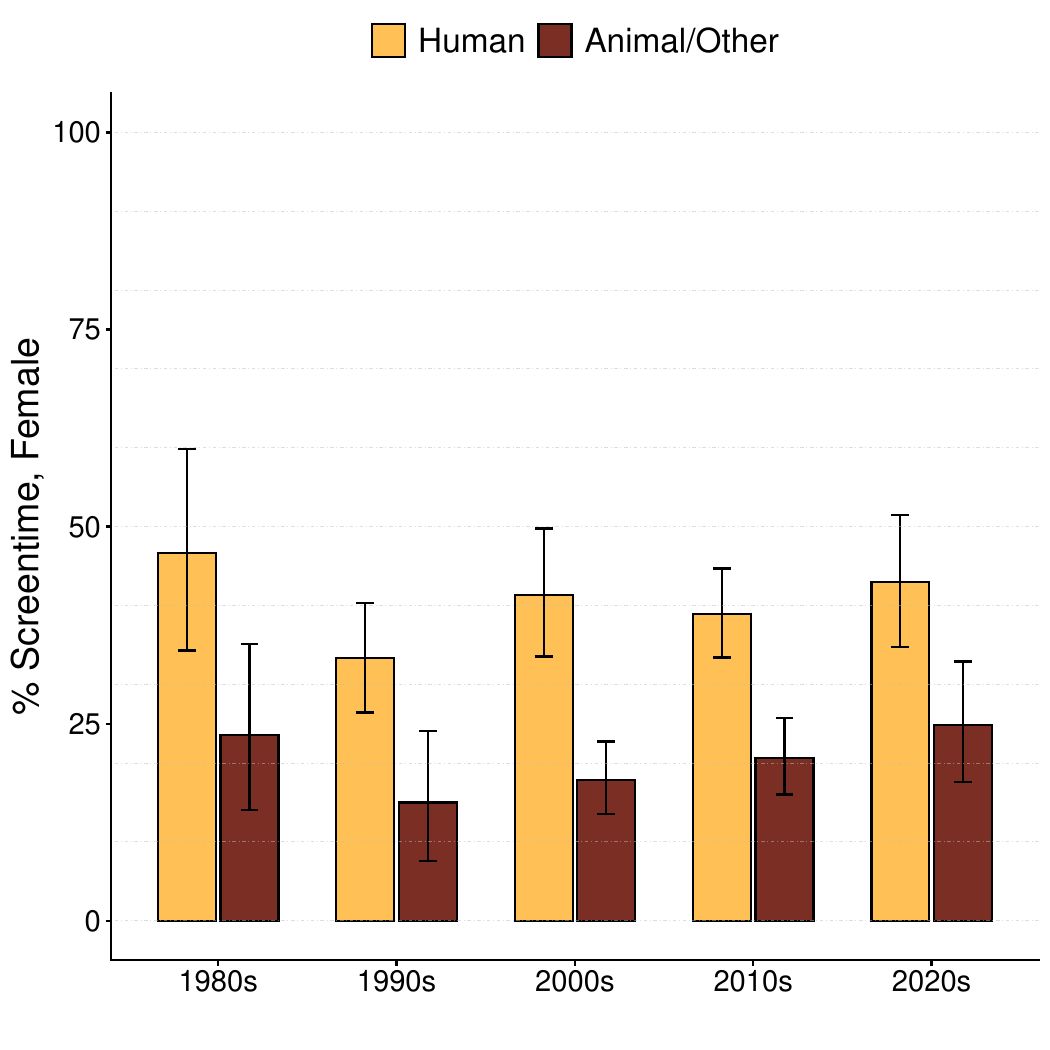}
    \caption{Animated human vs.\ animal/other}
    \label{fig:human-vs-animal}
  \end{subfigure}
  \caption{Screentime, female characters, 1980--2025.}
  \label{fig:screentime-comparison}
\end{figure}

First, do popular animated films reflect the same gender dynamics as contemporaneous live-action Hollywood films? Figure \ref{fig:total-vs-live-action} answers this by plotting the screentime of female animated characters in comparison to the screentime for women in live-action films over the period 1980--2025; live-action data is drawn directly from \textcite{bammanpnas}, and measures the complement to our work here (gender representation in the top 50 movies by US box office, excluding animation); while that work measures gender representation through 2022, we extend the live-action corpus with the same criteria  through 2025 for direct comparison with our animated collection.  At first glance, these two forms of film tell identical stories: male characters claim roughly 3 times more screentime than female characters, and we see a movement toward parity beginning in 2010.  In the strictest test, we can compare changing patterns of gender representation \emph{within} each franchise.  As figure \ref{fig:franchises} (appendix \ref{appdx:franchises}) illustrates, many franchises show increasing representation for female characters as each sequel is released.

But animated films portray a much wider variety of characters---not only humans, but also animals, machines, and fantastic creatures (as noted in \S\ref{sec:detection}). This is precisely where we might expect to see animation's imaginative visualization at work, and where we also might anticipate more of the creative liberty in relation to social norms and structures that Eisenstein associated with early animation. To explore this, we manually classified every character in our study into three categories: \emph{human} (and human-like, including rough human proportions), \emph{animal} (including dinosaurs) and \emph{other} (including robots, machines, monsters, etc.).  To assess the coherence of this classification, two annotators labeled the same 224 characters (one sampled from each movie) and see very high agreement (94\% raw agreement; Cohen's $\kappa$ = .90).

With this data, we can analyze the gender representation broken down by character category. Figure \ref{fig:human-vs-animal} presents these results, and we see a very different pattern emerging.  Human characters are far more often likely to be women in animated films than in comparable live-action ones, a finding that adds dimensionality to previous studies. Non-human characters (largely animals, which dominate the category) are even more disproportionately male than live-action casts. 

This masculinization of animal characters might be attributable to the ``default'' nature of maleness, hypothesized in Roman Jakobson's markedness theory \cite{jakobson1985selected}, which explores the marked (female) and unmarked (male) nature of terms in a binary, and explored in feminist interventions on the centering of the male at the expense of the female by Simone de Beauvoir \cite{de2015second} and Luce Irigaray \cite{irigaray1985sex}. Our finding serves as a quantitative complement to recent work of Jessica Birthisel \cite{birthisel2014body}, who finds lead characters in DreamWorks and Pixar productions between 2000 and 2010 to be overwhelmingly male and nonhuman. Masculinity in those films' anthropomorphized animated characters, she argues, is developed through bodily, sexual, and social coding. Conversely, Célia Jacquet \cite{jacquet2026challenging} argues that Disney's \textit{Beauty and the Beast }and \textit{Princess and the Frog} produce a new masculinity rooted in animals and the natural world that is their habitat. This argument is tempered, however, by both the small sample of films analyzed and the author's acknowledgment that, in both cases, the princes' animalization is reversed as the films conclude with a restoration of traditional social order. Our finding that female representation in human characters is increasing while representation in nonhuman characters is predominately male adds perspective to these qualitative studies and complicates the notion that increased representation equals decisive progress toward representational parity.   

\subsection{Protagonists and antagonists}

Animated films still strongly gender non-human characters as male (at rates even beyond the representation of men in live-action movies), but have consistently centered women and girls more often than live action. What roles do all of those characters play? As \textcite{davis2007good} notes, while early Disney heroines (such as Aurora in \emph{Sleeping Beauty}) are often depicted as passive, it is the villains (such as Maleficent) who are depicted as ``mature, powerful, and independent\ldots everything that their female victims are not'' (107).  While those protagonists have developed over the decades to become strong and independent on their own, it has come with a decline in the female antagonist: ``evil women are becoming an increasingly rare phenomenon'' (214).

We measure the comparative rates at which female and male characters play protagonist and antagonist roles in animated and live action films (across all characters---not only humans---since antagonists can often have non-human or boundary-crossing forms).  We draw on plot summaries on Wikipedia for providing a signal on these major roles; given a Wikipedia summary and the cast of characters---an IMDb cast list of names (for live-action films) and the character cast lists constructed in \S\ref{castlists} above---we prompt Gemini 3.1 Pro to identify all major and minor protagonists and antagonists mentioned---seeing this less as an interpretive challenge and more as an information extraction task (which character is described as the protagonist in the summary?) at which LLMs tend to excel.  We define a protagonist as ``the main character(s) whose goals/arc drive the story'' and antagonist as ``the character(s) who primarily oppose the protagonist(s) or drive the central conflict against them.'' (see appendix \ref{protprompt} for full prompt).  We evaluate the performance of Gemini at this task by comparing model outputs with human judgments on 262 characters from 30 sample movies (where a human selects the set of protagonists and antagonists in a film, and additionally marks them as \emph{major} or \emph{minor}). We generally find a reasonable F1 score in identifying protagonists and antagonists from the full cast (74.8\%), with the main cause of error coming from liminal judgments about minor characters; when restricted to \emph{major} characters alone, F1 rises to 93.7\%.

\begin{figure}[t]
    \centering
    \includegraphics[width=1.0\linewidth]{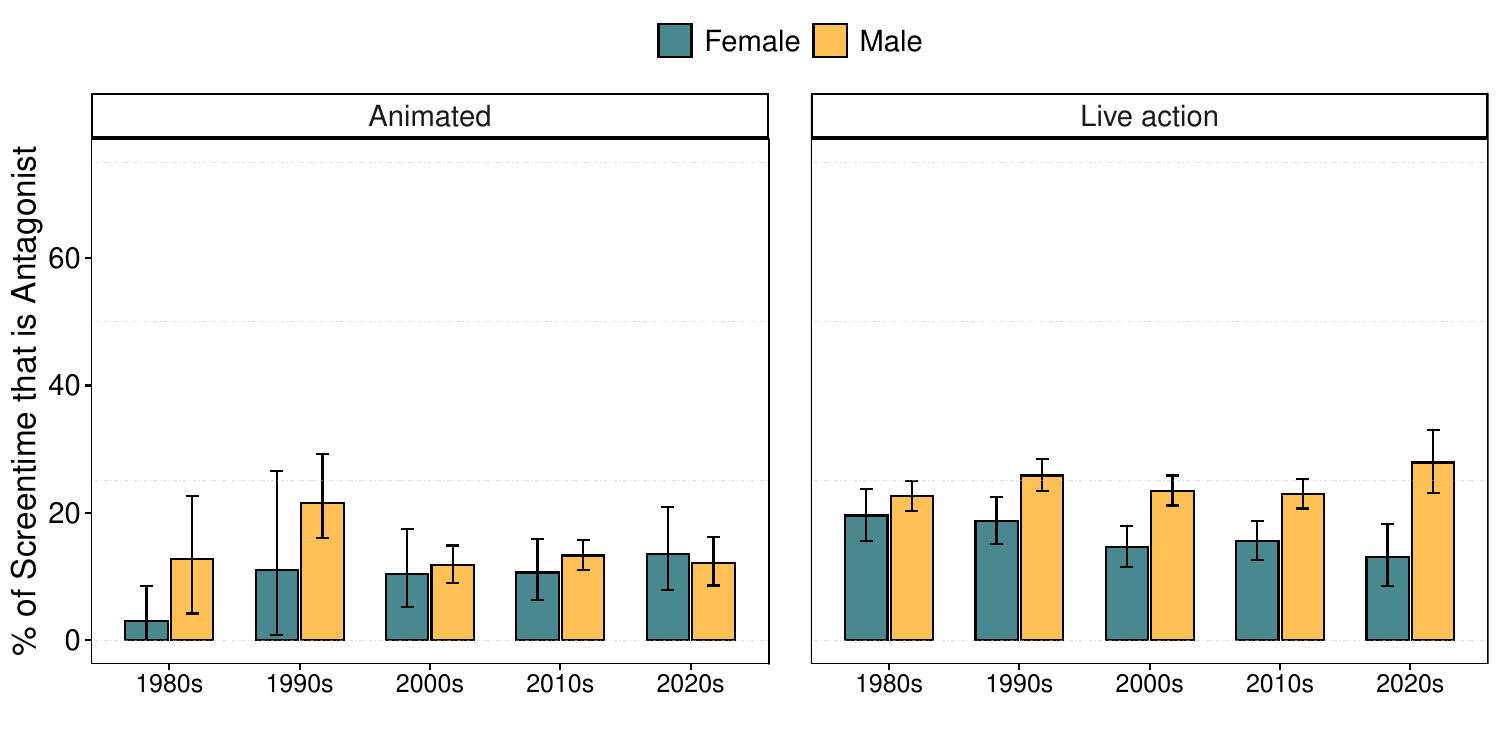}
    \caption{Antagonist screentime by gender, decade and animation status, with 95\% bootstrap confidence intervals (resampling at the level of movies).}
    \label{fig:prot}
\end{figure}

Using these predictions, we plot the comparative ratio of screentime that is dedicated to protagonists vs. antagonists by gender in figure \ref{fig:prot}, where each bar reflects the percentage (for each movie) of a given gender's combined protagonist and antagonist screentime given to the antagonist (i.e., only measuring screentime for protagonist/antagonist characters).  While the confidence intervals for animated films are large, we see no meaningful difference between female and male animated characters over this period of time;  in fact, it is only for live action films that we see a significant difference---for the entire 21st century, women have been cast far less often as antagonists than men.

\subsection{Sharing the screen}

In \emph{Good Girls and Wicked Witches}, Amy Davis argues that while women and girls in Disney movies have been represented with increasing complexity, independence and agency over time (over the period 1937--2005), a tendency has remained ``to present women as being largely in isolation from other women'' (228)---a trend that distinguishes Disney films from contemporaneous live action ones.

While Davis was arguing specifically about Disney, we take inspiration to examine this question within popular animation more broadly in order to put it in contrast to live-action film: do we see live action films broadly depicting women together on screen more so than contemporaneous animation?  To some degree, we can see this as a minimal prerequisite to the Bechdel test (assessing female agency by observing whether two named women are depicted as talking to each other about something other than a man).

\begin{figure}[t]
    \centering
    \includegraphics[width=0.6\linewidth]{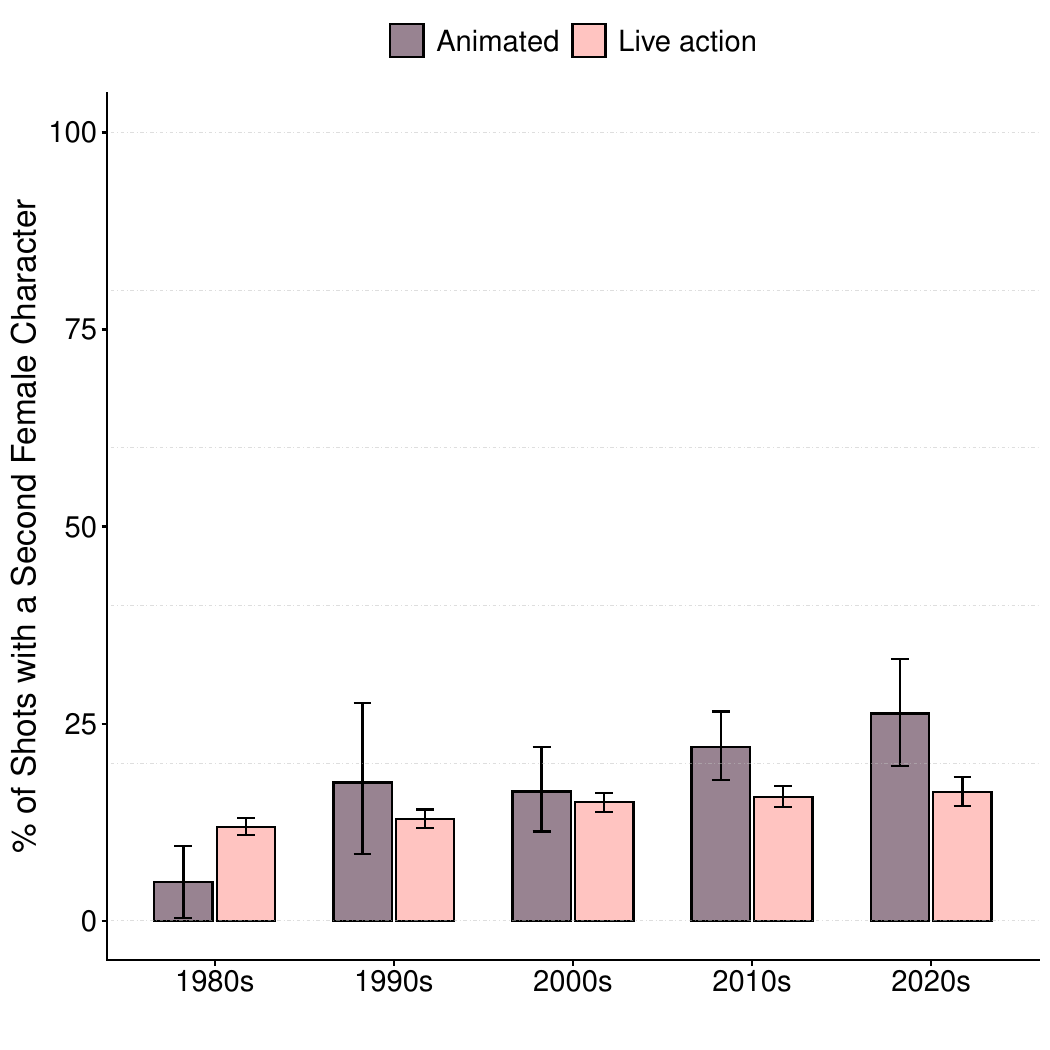}
    \caption{In shots that contain at least one female character, how many contain at least one other (in either that shot or the subsequent shot)?  Animated films generally show higher rates of co-presence among female characters than contemporaneous live-action ones.}
    \label{fig:female_cooccurrence_by_decade_fem}
\end{figure}

We test this by focusing exclusively on human characters, measuring the degree to which women and girls are depicted in isolation---i.e., what fraction of shots containing a female character depict \emph{at least} one other female character in that same shot or in the subsequent shot (to capture shot/reverse shot patterns). Figure \ref{fig:female_cooccurrence_by_decade_fem} illustrates the results.  While isolation may have been the norm for earlier Disney films, animated films in general from the 1990s on show greater co-occurrence (less isolation) than live action films released at the same time; women are both far more likely to be visible overall in animated films, and more likely to share the screen with other women.

\section{Conclusion}

In depicting an entirely created world, animated filmmaking has the capacity to not inherit the form of live-action cinema.  As our studies show, it both resists and yet embraces the androcentrism present there---representing women and girls (human characters) at greater rates than live action, but preserving even more extreme biases in animalization, where men give voice to non-human characters at roughly four times the rate of women. As our social and cultural lives have become increasingly intertwined with onscreen narratives that exert influence, previous studies have served to heighten awareness about these representational disparities; what this study offers to complement this existing body of research is both the capacity to identify larger trends---as animation has come to hold a larger and larger share of the viewing experience (roughly 20-25\% of the top box office revenues over the past ten years)---and a more fine-grained method to zoom in to any one of them (identifying who is visible, and when), and is especially useful when put in dialogue with the more traditional qualitative work done by scholars like Davis\ \cite{davis2007good}, Birthisel\ \cite{birthisel2014body}, and Jacquet\ \cite{jacquet2026challenging} (among others).

While this work draws attention to visibility, the role of a character (as protagonist/antagonist) and their co-occurrence with others, there is of course much left unexamined; as Davis notes, ``the real difference between the male and female characters is their level of activity (as opposed to passivity) within their stories'' \cite[109]{davis2007good}, and there is more work to be done to characterize the nature of that visibility (we should exercise caution when assuming that more screentime equals better screentime).  This becomes especially clear when our analysis is expanded to include gender in relation to non-human characters: the potential progress we noted in one area of representation may effectively be mitigated by the lack of parity elsewhere. 

Data and code to support this work---and enable other explorations of character in animated film---can be
found at \href{https://github.com/bamman-group/animated-gender-representation}{https://github.com/bamman-group/animated-gender-representation}; this includes trained models for character detection and recognition in animation, along with data produced by this pipeline on our collection of 224 films (including frame-level information on the location of detected characters in each film).

\section{Note on AI Usage}

We use Claude Code extensively for coding assistance in this work. This assistance results in computational scripts that are all publicly available for inspection and allow for reproducing the results reported here on iCartoonFace public data.  No AI was used in the writing of this paper, or in any manual annotations created.

\section*{Acknowledgments}

The research reported in this article was supported by the Humanities and AI Virtual Institute (HAVI),
a program of Schmidt Sciences.  Character clustering was aided by Sid Bell, Monriseth Escobar, Phoebe Marin, Julia Sayette, and Juni Singh, all student research fellows at Bowdoin College.
This work was made possible by the use of the Secure Research Data and Compute Platform at the University of California, Berkeley.

\clearpage

\printbibliography

@article{bochinski2017high,
  title={High-speed tracking-by-detection without using image information},
  author={Bochinski, Erik and Eiselein, Volker and Sikora, Thomas},
  journal={2017 14th IEEE international conference on advanced video and signal based surveillance (AVSS)},
  pages={1--6},
  year={2017},
  organization={IEEE}
}

@article{reagle2011gender,
  title={Gender bias in Wikipedia and Britannica},
  author={Reagle, Joseph and Rhue, Lauren},
  journal={International Journal of Communication},
  volume={5},
  pages={21--21},
  year={2011}
}

@online{walsh2025bears,
  author       = {Melanie Walsh and Russell Samora, Michelle Pera-McGhee, and Jan Diehm},
  title        = {Bears Will Be Boys: A Data Analysis of Animal Gender in Children's Books},
  year         = {2025},
  url          = {https://pudding.cool/2025/07/kids-books/},
  organization = {The Pudding},
}

@book{davis2007good,
  author    = {Amy M. Davis},
  title     = {Good Girls and Wicked Witches: Women in Disney's Feature
Animation},
  publisher = {Indiana University Press},
  year      = {2011},
}

@article{bamman2014unsupervised,
  title={Unsupervised discovery of biographical structure from text},
  author={Bamman, David and Smith, Noah A},
  journal={Transactions of the Association for Computational Linguistics},
  volume={2},
  pages={363--376},
  year={2014}
}

@unpublished{bamman2026evaluating,
  author = {David Bamman and Kent K. Chang and Allison Cooper and Juishan Hsu and Reina Kushihashi and Madison Mar and Arnav Podichetty and Rachael Samberg and Ipek Nil Sancak and Yuhan Shao},
  title = {Evaluating Multimodal Narrative Understanding of Popular Hollywood Films},
  year = {2026},
  note = {Preprint}
}

@misc{guo2019insightface,
  title={Insightface: {2D} and {3D} face analysis project},
  author={Guo, Jia and Deng, Jiankang},
  year={2019},
  howpublished={\url{https://github.com/deepinsight/insightface}}

}

@article{gui2025character,
          title={Character-Centric Understanding of Animated Movies},
          author={Gui, Zhongrui and Xie, Junyu and Han, Tengda and Xie, Weidi and Zisserman, Andrew},
          journal={arXiv preprint arXiv:2509.12204},
          year={2025}
        }

@inproceedings{deng2019arcface,
  title={Arcface: Additive angular margin loss for deep face recognition},
  author={Deng, Jiankang and Guo, Jia and Xue, Niannan and Zafeiriou, Stefanos},
  booktitle={Proceedings of the IEEE/CVF conference on computer vision and pattern recognition},
  pages={4690--4699},
  year={2019}
}

@article{oquab2023dinov2,
  title={Dinov2: Learning robust visual features without supervision},
  author={Oquab, Maxime and Darcet, Timoth{\'e}e and Moutakanni, Th{\'e}o and Vo, Huy and Szafraniec, Marc and Khalidov, Vasil and Fernandez, Pierre and Haziza, Daniel and Massa, Francisco and El-Nouby, Alaaeldin and others},
  journal={arXiv preprint arXiv:2304.07193},
  year={2023}
}

@article{jocher2026ultralytics,
  title={Ultralytics yolo26: Unified real-time end-to-end vision models},
  author={Jocher, Glenn and Qiu, Jing and Liu, Mengyu and Lyu, Shuai and Akyon, Fatih Cagatay and Kalfaoglu, Muhammet Esat},
  journal={arXiv preprint arXiv:2606.03748},
  year={2026}
}

@inproceedings{yang2016wider,
  title={Wider face: A face detection benchmark},
  author={Yang, Shuo and Luo, Ping and Loy, Chen-Change and Tang, Xiaoou},
  booktitle={Proceedings of the IEEE conference on computer vision and pattern recognition},
  pages={5525--5533},
  year={2016}
}

@article{bochkovskiy2020yolov4,
  title={Yolov4: Optimal speed and accuracy of object detection},
  author={Bochkovskiy, Alexey and Wang, Chien-Yao and Liao, Hong-Yuan Mark},
  journal={arXiv preprint arXiv:2004.10934},
  year={2020}
}

@article{sapkota2026yoloe,
  title={YOLOE-26: Integrating YOLO26 with YOLOE for Real-Time Open-Vocabulary Instance Segmentation},
  author={Sapkota, Ranjan and Karkee, Manoj},
  journal={arXiv preprint arXiv:2602.00168},
  year={2026}
}

@article{wu2023yunet,
  title={Yunet: A tiny millisecond-level face detector},
  author={Wu, Wei and Peng, Hanyang and Yu, Shiqi},
  journal={Machine Intelligence Research},
  volume={20},
  number={5},
  pages={656--665},
  year={2023},
  publisher={Springer}
}

@misc{lv2024rtdetrv2improvedbaselinebagoffreebies,
      title={RT-DETRv2: Improved Baseline with Bag-of-Freebies for Real-Time Detection Transformer}, 
      author={Wenyu Lv and Yian Zhao and Qinyao Chang and Kui Huang and Guanzhong Wang and Yi Liu},
      year={2024},
      eprint={2407.17140},
      archivePrefix={arXiv},
      primaryClass={cs.CV},
      url={https://arxiv.org/abs/2407.17140}, 
}

@inproceedings{lin2014microsoft,
  title={Microsoft coco: Common objects in context},
  author={Lin, Tsung-Yi and Maire, Michael and Belongie, Serge and Hays, James and Perona, Pietro and Ramanan, Deva and Doll{\'a}r, Piotr and Zitnick, C Lawrence},
  booktitle={European conference on computer vision},
  pages={740--755},
  year={2014},
  organization={Springer}
}

@inproceedings{zheng2020cartoon,
title={Cartoon Face Recognition: A Benchmark Dataset},
author={Zheng, Yi and Zhao, Yifan and Ren, Mengyuan and Yan, He and Lu, Xiangju and Liu, Junhui and Li, Jia},
booktitle={Proceedings of the 28th ACM International Conference on Multimedia},
pages={2264--2272},
year={2020}
}

@inproceedings{deng2020retinaface,
  title={Retinaface: Single-shot multi-level face localisation in the wild},
  author={Deng, Jiankang and Guo, Jia and Ververas, Evangelos and Kotsia, Irene and Zafeiriou, Stefanos},
  booktitle={Proceedings of the IEEE/CVF conference on computer vision and pattern recognition},
  pages={5203--5212},
  year={2020}
}

@article{zhu2020tinaface,
  title={Tinaface: Strong but simple baseline for face detection},
  author={Zhu, Yanjia and Cai, Hongxiang and Zhang, Shuhan and Wang, Chenhao and Xiong, Yichao},
  journal={arXiv preprint arXiv:2011.13183},
  year={2020}
}

@article{soucek2020transnetv2,
    title={TransNet V2: An effective deep network architecture for fast shot transition detection},
    author={Sou{\v{c}}ek, Tom{\'a}{\v{s}} and Loko{\v{c}}, Jakub},
    year={2020},
    journal={arXiv preprint arXiv:2008.04838},
}

@article{
bammanpnas,
author = {David Bamman  and Rachael Samberg  and Richard Jean So  and Naitian Zhou },
title = {Measuring diversity in Hollywood through the large-scale computational analysis of film},
journal = {Proceedings of the National Academy of Sciences},
volume = {121},
number = {46},
pages = {e2409770121},
year = {2024},
doi = {10.1073/pnas.2409770121},
URL = {https://www.pnas.org/doi/abs/10.1073/pnas.2409770121},
}

@techreport{geena2019seejane,
  author       = {{Geena Davis Institute on Gender in Media}},
  title        = {See Jane 2019: An Analysis of Representations in Film and Television},
  institution  = {Geena Davis Institute on Gender in Media},
  year         = {2019},
  url          = {https://geenadavisinstitute.org/wp-content/uploads/2024/01/see-jane-2019-full-report.pdf}
}

@techreport{teran2024seejane,
  author      = {Terán, L. and Conroy, M.},
  title       = {See Jane 2024: How Has On-Screen Representation in Children's Television Changed from 2018 to 2023?},
  institution = {Geena Davis Institute},
  year        = {2024},
  url         = {https://geenadavisinstitute.org/wp-content/uploads/2024/09/GDI-September-2024_See-Jane-2024_How-Has-On-Screen-Representation-in-Childrens-Television-Changed-from-2018-to-2023.pdf}
}

@techreport{geena2019benchmark,
  author       = {{Geena Davis Institute on Gender in Media}},
  title        = {The Geena Benchmark Report: 2007--2017},
  institution  = {Geena Davis Institute on Gender in Media},
  year         = {2019},
  url          = {https://geenadavisinstitute.org/wp-content/uploads/2024/01/geena-benchmark-report-2007-2017-2-12-19.pdf},
}

@article{raju2023cohort,
  title={A cohort study of the diversity in animated films from 1937 to 2021: in a world less enchanted can we be more encanto?},
  author={Raju, Suneil A and Sanders, Samira R and Bolton-Raju, Kathryn S and Bowker-Howell, Freya J and Hall, Lara R and Newton, Millie and Neill, Gary S and Holland, William J and Howford, Katie L and Bolton, Emma V and others},
  journal={Cureus},
  volume={15},
  number={8},
  year={2023},
  publisher={Cureus}
}

@techreport{bls_atus_2024,
  author      = {{U.S. Bureau of Labor Statistics}},
  title       = {American Time Use Survey -- 2025 Results},
  institution = {U.S. Bureau of Labor Statistics},
  year        = {2025},
  type        = {Technical Report},
  url         = {https://www.bls.gov/news.release/atus.htm}
}

@misc{horst2015open,
  title={An open book: What and how young children learn from picture and story books},
  author={Horst, Jessica S and Houston-Price, Carmel},
  journal={Frontiers in psychology},
  volume={6},
  pages={1719},
  year={2015},
  publisher={Frontiers Media SA}
}

@misc{abad2013storybooks,
  title={Do storybooks really break children's gender stereotypes?},
  author={Abad, Carla and Pruden, Shannon M},
  year={2013},
  publisher={Frontiers Media SA}
}

@article{underwood2018transformation,
  title={The transformation of gender in English-language fiction},
  author={Underwood, Ted and Bamman, David and Lee, Sabrina},
  journal={Journal of Cultural Analytics},
  volume={3},
  number={2},
  year={2018},
  publisher={Center for Digital Humanities, Princeton University}
}

@article{casey2021sixty,
    doi = {10.1371/journal.pone.0260566},
    author = {Casey, Kennedy AND Novick, Kylee AND Lourenco, Stella F.},
    journal = {PLOS ONE},
    publisher = {Public Library of Science},
    title = {Sixty years of gender representation in children’s books: Conditions associated with overrepresentation of male versus female protagonists},
    year = {2021},
    month = {12},
    volume = {16},
    pages = {1-19},
    number = {12},

}

@book{EisensteinSergei1988EoD,
publisher = {Methuen},
isbn = {0413196402},
year = {1988},
title = {Eisenstein on Disney / edited by Jay Leyda ; translated by Alan Upchurch ; introduced by Naum Kleiman.},
address = {London},
author = {Eisenstein, Sergei},
}

@book{dobson2018animation,
  title={The animation studies reader},
  author={Dobson, Nichola and Roe, Annabelle Honess and Ratelle, Amy and Ruddell, Caroline},
  year={2018},
  publisher={Bloomsbury Academic}
}

@book{loock2024hollywood,
  title={Hollywood remaking: How film remakes, sequels, and franchises shape industry and culture},
  author={Loock, Kathleen},
  year={2024},
  publisher={University of California Press Oakland, CA}
}

@article{jacquet2026challenging,
  title={Challenging Hierarchies Through Animality: Interspecies and Gender Relations in Disney’s Beauty and the Beast and The Princess and the Frog},
  author={Jacquet, C{\'e}lia},
  journal={Animals},
  volume={16},
  number={7},
  pages={1055},
  year={2026},
  publisher={MDPI}
}

@book{jakobson1985selected,
  title={Selected writings},
  author={Jakobson, Roman},
  year={1985},
  publisher={Walter de Gruyter}
}

@book{de2015second,
  title={The Second Sex (Vintage Feminism Short Edition)},
  author={De Beauvoir, Simone},
  year={2015},
  publisher={Random House}
}

@book{irigaray1985sex,
  title={This sex which is not one},
  author={Irigaray, Luce},
  year={1985},
  publisher={Cornell university press}
}

@article{birthisel2014body,
  title={How body, heterosexuality and patriarchal entanglements mark non-human characters as male in CGI-animated children's films},
  author={Birthisel, Jessica},
  journal={Journal of Children and Media},
  volume={8},
  number={4},
  pages={336--352},
  year={2014},
  publisher={Taylor \& Francis}
}

@book{giroux2001mouse,
  title={The mouse that roared: Disney and the end of innocence},
  author={Giroux, Henry A},
  year={2001},
  publisher={Bloomsbury Publishing PLC}
}

@misc{bandura1986social,
  title={Social foundations of thought and action: A social cognitive theory},
  author={Bandura, Albert},
  year={1986},
  publisher={Prentice-hall}
}

@incollection{gerbner2002growing,
  title={Growing up with television: Cultivation processes},
  author={Gerbner, George and Gross, Larry and Morgan, Michael and Signorielli, Nancy and Shanahan, James},
  booktitle={Media effects},
  pages={53--78},
  year={2002},
  publisher={Routledge}
}

@book{tuchman1978hearth,
  title     = {Hearth and Home: Images of Women in the Mass Media},
  editor    = {Tuchman, Gaye and Daniels, Arlene Kaplan and Ben{\'e}t, James},
  year      = {1978},
  publisher = {Oxford University Press},
  address   = {New York}
}

@book{althusser2006lenin,
  title={Lenin and philosophy and other essays},
  author={Althusser, Louis},
  year={2006},
  publisher={Aakar Books}
}

@article{piluso2023above,
  title={Above the heteronormative narrative: Looking up the place of Disney’s villains},
  author={Piluso, Francesco},
  journal={Semiotica},
  volume={2023},
  number={255},
  pages={131--148},
  year={2023},
  publisher={De Gruyter}
}

@book{bem1993lenses,
  title={The lenses of gender},
  author={Bem, Sandra Lipsitz},
  volume={40},
  year={1993},
  publisher={New Haven, CT: Yale University Press}
}

\clearpage

\appendix

\section{Author Contributions}

\begin{itemize}

 \item Animated character detection/recognition annotation: RAR, RK, MM

 \item Character clustering annotation: RAR, RK, MM

 \item Character category (human, animal, other) annotation: DB, RK

 \item Character protagonist/antagonist annotation: RAR, DB

 \item Statistical analysis: DB

 \item Computational methodology: DB

 \item Validation: DB

 \item Writing and theoretical framing: DB, AC

\end{itemize}

\section{Detection} \label{appdx:detect}

\begin{table}[h!]
\centering
\begin{tabular}{lrrrr}
\toprule
Model & Train WF & Train ICF & Train WF+ICF & im/s \\
\midrule
RetinaFace-R50 & 19.17 {\scriptsize [18.35, 20.06]} & 89.60 {\scriptsize [88.96, 90.23]} & 89.73 {\scriptsize [89.12, 90.34]} & 58.5 \\
\quad + TTA & 26.44 {\scriptsize [25.44, 27.51]} & 89.57 {\scriptsize [88.96, 90.17]} & 89.75 {\scriptsize [89.16, 90.31]} & 3.9 \\
YOLO26-L & 33.06 {\scriptsize [32.00, 34.11]} & 90.61 {\scriptsize [90.02, 91.17]} & 91.13 {\scriptsize [90.54, 91.71]} & 111.2 \\
\quad + TTA & 39.15 {\scriptsize [38.14, 40.32]} & 91.48 {\scriptsize [90.94, 92.05]} & 91.76 {\scriptsize [91.18, 92.29]} & 13.9 \\
RT-DETR-L & 39.76 {\scriptsize [38.61, 40.91]} & 91.92 {\scriptsize [91.37, 92.45]} & 91.69 {\scriptsize [91.16, 92.24]} & 86.4 \\
\quad + TTA & 42.19 {\scriptsize [41.08, 43.33]} & 91.53 {\scriptsize [91.03, 92.05]} & 91.68 {\scriptsize [91.16, 92.16]} & 8.5 \\
YuNet-N & 26.39 {\scriptsize [25.54, 27.33]} & 80.01 {\scriptsize [79.13, 80.85]} & 78.74 {\scriptsize [77.86, 79.61]} & 142.9 \\
\quad + TTA & 33.41 {\scriptsize [32.43, 34.53]} & 80.22 {\scriptsize [79.42, 81.04]} & 78.48 {\scriptsize [77.70, 79.33]} & 19.2 \\
\midrule
YOLOE-26L (text-prompted) & \multicolumn{3}{c}{14.03 {\scriptsize [13.46, 14.68]}} & 95.7 \\
\quad + TTA & \multicolumn{3}{c}{20.30 {\scriptsize [19.53, 21.14]}} & 6.9 \\
\bottomrule
\end{tabular}
\caption{AP@0.5 (95\% bootstrap CI) on iCartoonFace, plus inference throughput (images/second).}
\label{tab:icartoon_test}
\end{table}

\section{Recognition}\label{appdx:recog}

\begin{table}[h!]
\centering
\resizebox{\linewidth}{!}{%
\begin{tabular}{lccccr}
\toprule
Model & Base & FT & I+T/Train & I+T & im/s \\
\midrule
buffalo\_l & 0.0901 {\tiny [0.0851, 0.0955]} & 0.7148 {\tiny [0.7044, 0.7259]} & 0.6210 {\tiny [0.6097, 0.6319]} & 0.6180 {\tiny [0.6063, 0.6288]} & 312.9 \\
\quad + 25\% padding & 0.1670 {\tiny [0.1594, 0.1749]} & 0.7929 {\tiny [0.7834, 0.8026]} & 0.7263 {\tiny [0.7164, 0.7371]} & 0.7174 {\tiny [0.7067, 0.7282]} & 312.9 \\
DINOv2 vitb14 & 0.1705 {\tiny [0.1609, 0.1796]} & 0.8234 {\tiny [0.8142, 0.8330]} & 0.7591 {\tiny [0.7484, 0.7693]} & 0.7411 {\tiny [0.7305, 0.7516]} & 463.2 \\
\quad + 25\% padding & 0.1845 {\tiny [0.1739, 0.1940]} & 0.8669 {\tiny [0.8585, 0.8750]} & 0.8315 {\tiny [0.8223, 0.8401]} & 0.7976 {\tiny [0.7879, 0.8074]} & 463.2 \\
DINOv2 vitl14 & 0.1870 {\tiny [0.1771, 0.1967]} & 0.8434 {\tiny [0.8344, 0.8522]} & 0.7730 {\tiny [0.7624, 0.7833]} & 0.7280 {\tiny [0.7173, 0.7388]} & 155.1 \\
\quad + 25\% padding & 0.1957 {\tiny [0.1842, 0.2061]} & 0.8840 {\tiny [0.8759, 0.8922]} & 0.8282 {\tiny [0.8190, 0.8370]} & 0.7895 {\tiny [0.7796, 0.7994]} & 155.1 \\
\bottomrule
\end{tabular}
}
\caption{Rank@1 identification accuracy (95\% bootstrap CI) on iCartoonFace, crop 0\% and crop 25\% (indented), plus inference throughput (images/second).}
\label{tab:recognition_variants_icartoonface}
\end{table}

\clearpage
\section{Manual clustering}\label{appdx:clustering}

\begin{figure}[htbp]
    \centering
    \includegraphics[width=1\linewidth]{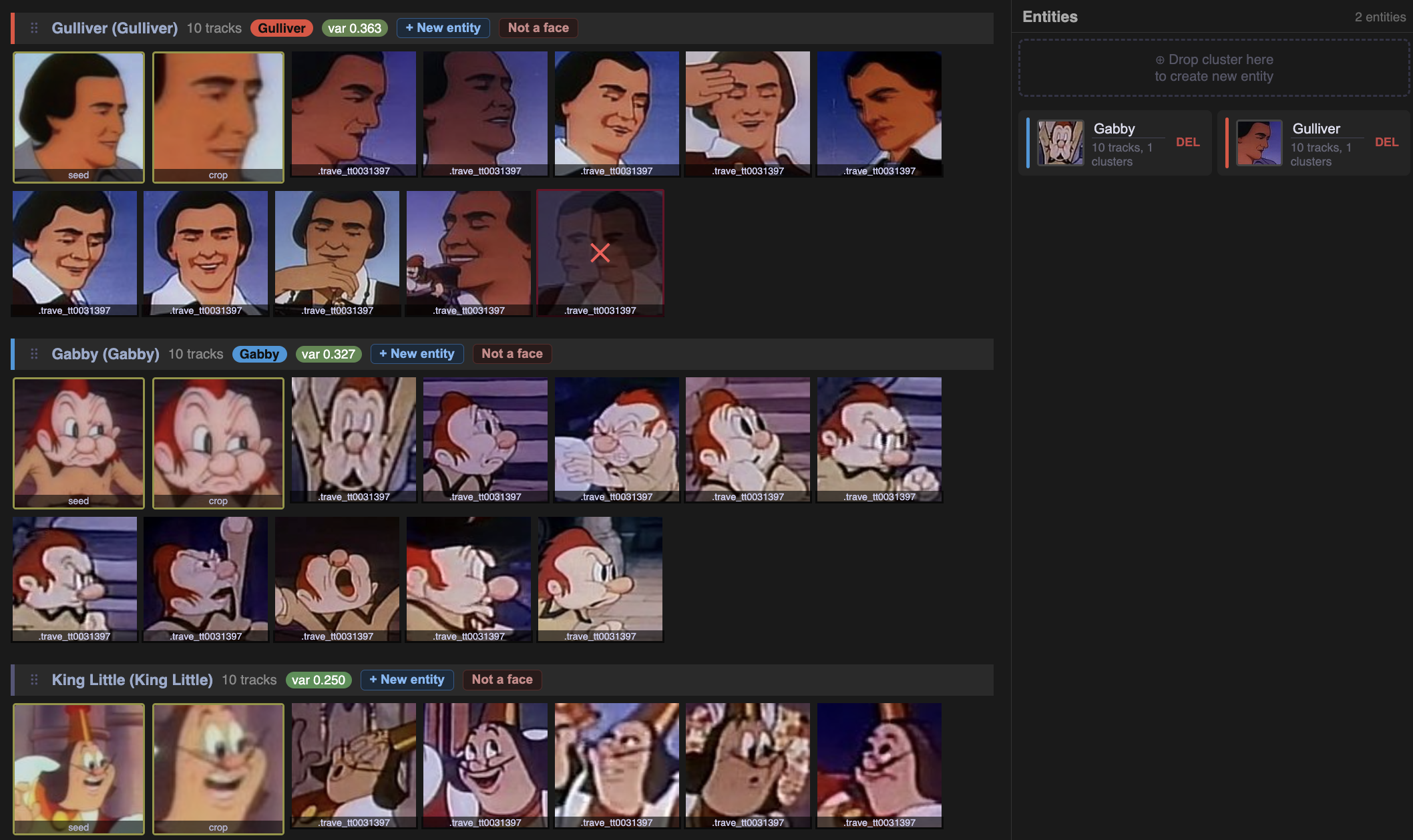}
    \caption{Annotation interface}
    \label{fig:annotation}
\end{figure}

Figure \ref{fig:annotation} illustrates the manual clustering annotation interface, where an annotator considers the candidate clusters on the left panel (each with 12 images) and creates a cast list of entities in the right panel (two entities have been created at this point in the annotation process).  One image from the first cluster (Gulliver) has been deselected.

\clearpage
\section{Franchises}\label{appdx:franchises}

\begin{figure}[htbp!]
    \centering
    \includegraphics[width=1\linewidth]{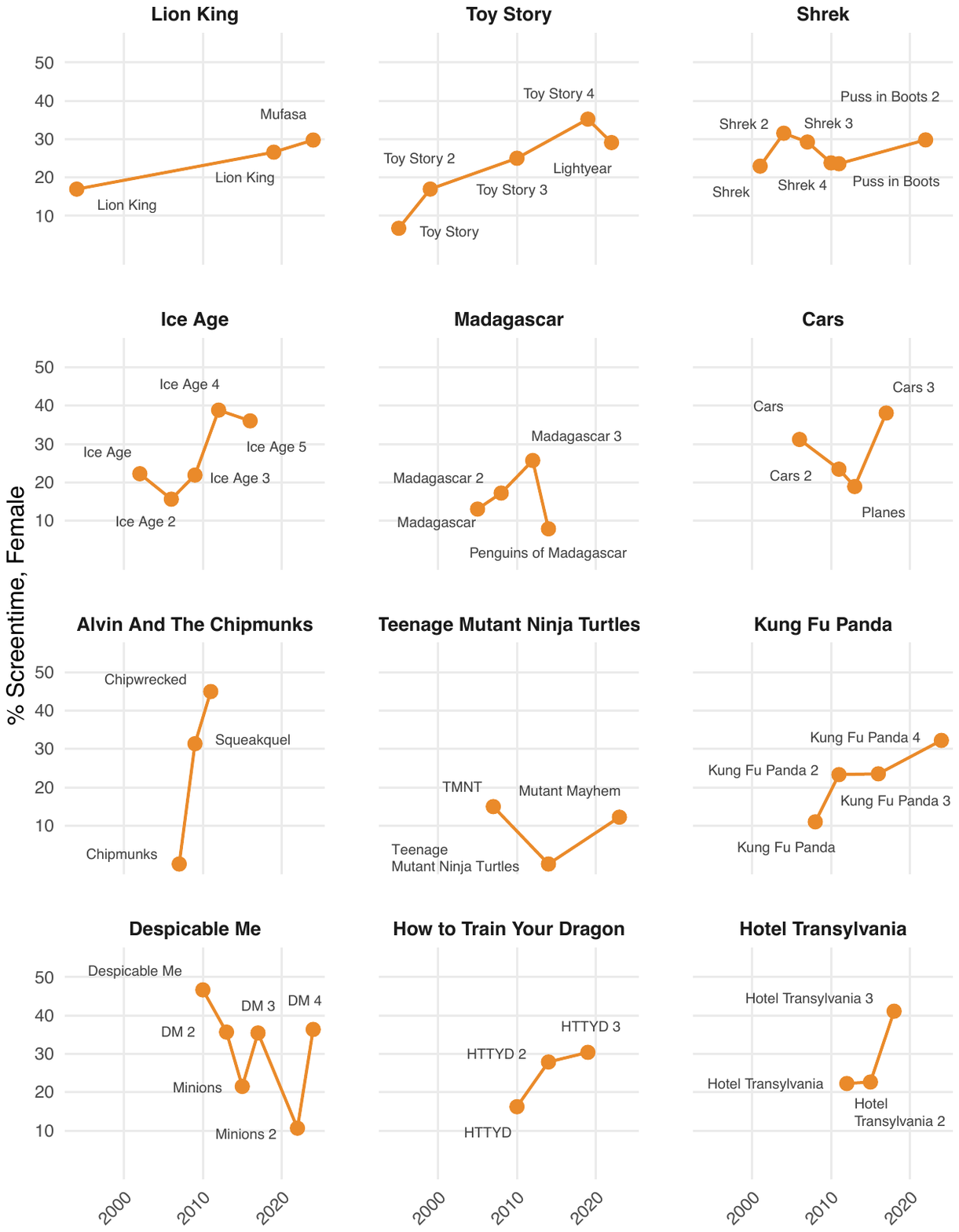}
    \caption{Representation for female characters across franchises, for all franchises with at least 3 movies in our collection.}
    \label{fig:franchises}
\end{figure}

\section{Prompt for identifying protagonists/antagonists}\label{protprompt}

Prompt used to identify protagonists and antagonists from a Wikipedia plot
summary:

\begin{quote}
\small
\begin{verbatim}
You are a film narrative analyst. You will be given a movie's title, a plot
summary, and the complete cast list for that movie, each entry given as
"character name (actor imdb id)". Identify which of those entries (using them
EXACTLY as given in the list, verbatim, including the "(actor imdb id)" part)
are the movie's protagonist(s) and antagonist(s), based only on the plot
summary provided.

- protagonist(s): the main character(s) whose goals and character arc drive
  the story.

- antagonist(s): the character(s) who primarily oppose the protagonist(s) or
  drive the central conflict against them.

For each character you name, also judge whether they are a "major" or "minor"
character. A major character is central to the plot summary and plays a
substantial role in driving events. A minor character serves the
protagonist/antagonist function only briefly, peripherally, or during a limited
part of the story (e.g., a secondary villain who appears in one act, or a
co-lead with a much smaller role than the primary protagonist).

Only choose entries from the given cast list. Do not invent names, and do not
include a character who is not in the list, even if the plot summary mentions
them. If the plot summary does not provide enough information to identify a
role with confidence, leave the corresponding list empty rather than guessing.
\end{verbatim}
\end{quote}

\end{document}